\documentclass[conference]{IEEEtran}
\IEEEoverridecommandlockouts
\usepackage{cite}
\usepackage{amsmath,amssymb,amsfonts}
\usepackage{algorithmic}
\usepackage{graphicx}
\usepackage{textcomp}
\usepackage{xcolor}
\usepackage{xcolor,colortbl}
\usepackage{threeparttable}
\definecolor{Gray}{gray}{0.85}
\def\BibTeX{{\rm B\kern-.05em{\sc i\kern-.025em b}\kern-.08em
    T\kern-.1667em\lower.7ex\hbox{E}\kern-.125emX}}
\begin{document}

\title{Machine-Learning-Based Classification of GPS Signal Reception Conditions Using a Dual-Polarized Antenna in Urban Areas\\
\thanks{This research was supported by the Unmanned Vehicles Core Technology Research and Development Program through the National Research Foundation of Korea (NRF) and the Unmanned Vehicle Advanced Research Center (UVARC) funded by the Ministry of Science and ICT, Republic of Korea (2020M3C1C1A01086407).
This research was also conducted as a part of the project titled ``Development of integrated R-Mode navigation system [PMS4440]'' funded by the Ministry of Oceans and Fisheries, Republic of Korea (20200450).}
}

\makeatletter
\newcommand{\linebreakand}{%
  \end{@IEEEauthorhalign}
  \hfill\mbox{}\par
  \mbox{}\hfill\begin{@IEEEauthorhalign}
}
\makeatother

\author{\IEEEauthorblockN{Sanghyun Kim}
\IEEEauthorblockA{
\textit{School of Integrated Technology} \\
\textit{Yonsei University}\\
Incheon, Korea \\
sanghyun.kim@yonsei.ac.kr}
\and
\IEEEauthorblockN{Jiwon Seo}
\IEEEauthorblockA{
\textit{School of Integrated Technology} \\
\textit{Yonsei University}\\
Incheon, Korea \\
jiwon.seo@yonsei.ac.kr}
}

\maketitle

\begin{abstract}
In urban areas, dense buildings frequently block and reflect global positioning system (GPS) signals, resulting in the reception of a few visible satellites with many multipath signals. This is a significant problem that results in unreliable positioning in urban areas. If a signal reception condition from a certain satellite can be detected, the positioning performance can be improved by excluding or de-weighting the multipath contaminated satellite signal. Thus, we developed a machine-learning-based method of classifying GPS signal reception conditions using a dual-polarized antenna. 
We employed a decision tree algorithm for classification using three features, one of which can be obtained only from a dual-polarized antenna. 
A machine-learning model was trained using GPS signals collected from various locations. 
When the features extracted from the GPS raw signal are input, the generated machine-learning model outputs one of the three signal reception conditions: non-line-of-sight (NLOS) only, line-of-sight (LOS) only, or LOS+NLOS. 
Multiple testing datasets were used to analyze the classification accuracy, which was then compared with an existing method using dual single-polarized antennas. 
Consequently, when the testing dataset was collected at different locations from the training dataset, a classification accuracy of 64.47\% was obtained, which was slightly higher than the accuracy of the existing method using dual single-polarized antennas. 
Therefore, the dual-polarized antenna solution is more beneficial than the dual single-polarized antenna solution because it has a more compact form factor and its performance is similar to that of the other solution.
\end{abstract}


\section{Introduction}
As new modes of urban transportation, such as urban air mobility, are emerging, the use of global navigation satellite systems (GNSS), particularly the global positioning system (GPS), in urban areas is expected to increase significantly \cite{Causa21, Kim20:Motion, Yoon20, Sun21:Markov, Lee22:Optimal, Sun20:Performance, Jia21:Ground, Lee20:Integrity, Lee19:Safety}. 
However, in urban areas, the multipath effect of GNSS signals can result in unreliable positioning results \cite{Lee22:Urban, Lee20:Preliminary, Kim22:First, Lee22:SFOL, Rhee21:Enhanced, Park20:Effects, MacGougan02, Shen20, Lee20, Lee23:Performance}. 
This is because of two major problems caused by dense buildings in urban areas. First, line-of-sight (LOS) signals can be blocked by buildings, which dramatically reduces the number of visible satellites. Second, GNSS signals can be reflected by buildings, thereby increasing the reception of non-line-of-sight (NLOS) signals. 
The first problem can be improved with the aid of multiple GNSS constellations, but the second one remains a significant challenge.

Signals from a certain GNSS satellite can be received under three different conditions in an urban area: only the LOS signal is received (i.e., LOS-only condition), only the NLOS signal is received (i.e., NLOS-only condition), or both LOS and NLOS signals are received simultaneously (i.e., LOS+NLOS condition). Note that the NLOS-only and LOS+NLOS conditions are considered to be different phenomena because they can produce different range errors \cite{Zhu18}. 
In the LOS+NLOS condition, the range error is bounded because a reflected signal with more than 1.5 chip delay is suppressed in the correlation process \cite{Enge1994:global}, whereas in the NLOS-only condition, the range error is unbounded. 
If multipath conditions (i.e., NLOS-only or LOS+NLOS conditions) are detected, the positioning performance can be improved by excluding or de-weighting the multipath contaminated signals \cite{Kubo20, Suzuki21}. Therefore, various multipath detection methods have been actively studied \cite{Closas11, Massarweh20, Lee22:Performance}.

One method uses a fish-eye camera to identify satellites in a sky plot covered by buildings as NLOS-only conditions \cite{Bai20}. 
A method for creating a sky plot utilizing a 3D city model was also proposed \cite{Wang12}. 
However, these methods are limited in that they can classify the signals into only two conditions: NLOS-only and the other (i.e., LOS-only or LOS+NLOS) conditions. 
Another method uses a 3D city model and ray-tracing technique to classify all three (i.e., LOS-only, LOS+NLOS, and NLOS-only) conditions \cite{Miura13}. 
Although this method can classify satellite signals under the three conditions, it has a limitation in that it is computationally intensive and it needs significant work to establish and load a 3D city model \cite{Zhu18}.

In addition, there are several statistical methods available for detecting multipath at the measurement level.
One such method detects multipaths by comparing measurements of the carrier-to-noise-density ratio ($C/N_0$) at three frequencies \cite{Strode16}. This is based on the principle that the difference in $C/N_0$ between the frequencies changes in a multipath environment. Similarly, the $C/N_0$ difference between the right-hand-circularly-polarized (RHCP) and left-hand-circularly-polarized (LHCP) signals of the dual-polarized antenna have been used \cite{Groves10, Kim21:GPS}. 
When a GNSS signal is reflected, its polarization changes from RHCP to LHCP. 
Accordingly, multipath conditions can be detected by observing the $C/N_0$ difference between the RHCP and LHCP signals, which is affected by the multipath environment. 
Other methods use measurements such as the code-minus-carrier (CMC) and Doppler shift \cite{Caamano20, Xu19}.
Furthermore, emerging studies are using machine-learning approaches to develop signal reception condition classifiers using GNSS measurements as features of machine learning \cite{Hsu17, Suzuki20, Sun21}.

\begin{figure}
  \centering
  \includegraphics[width=0.8\linewidth]{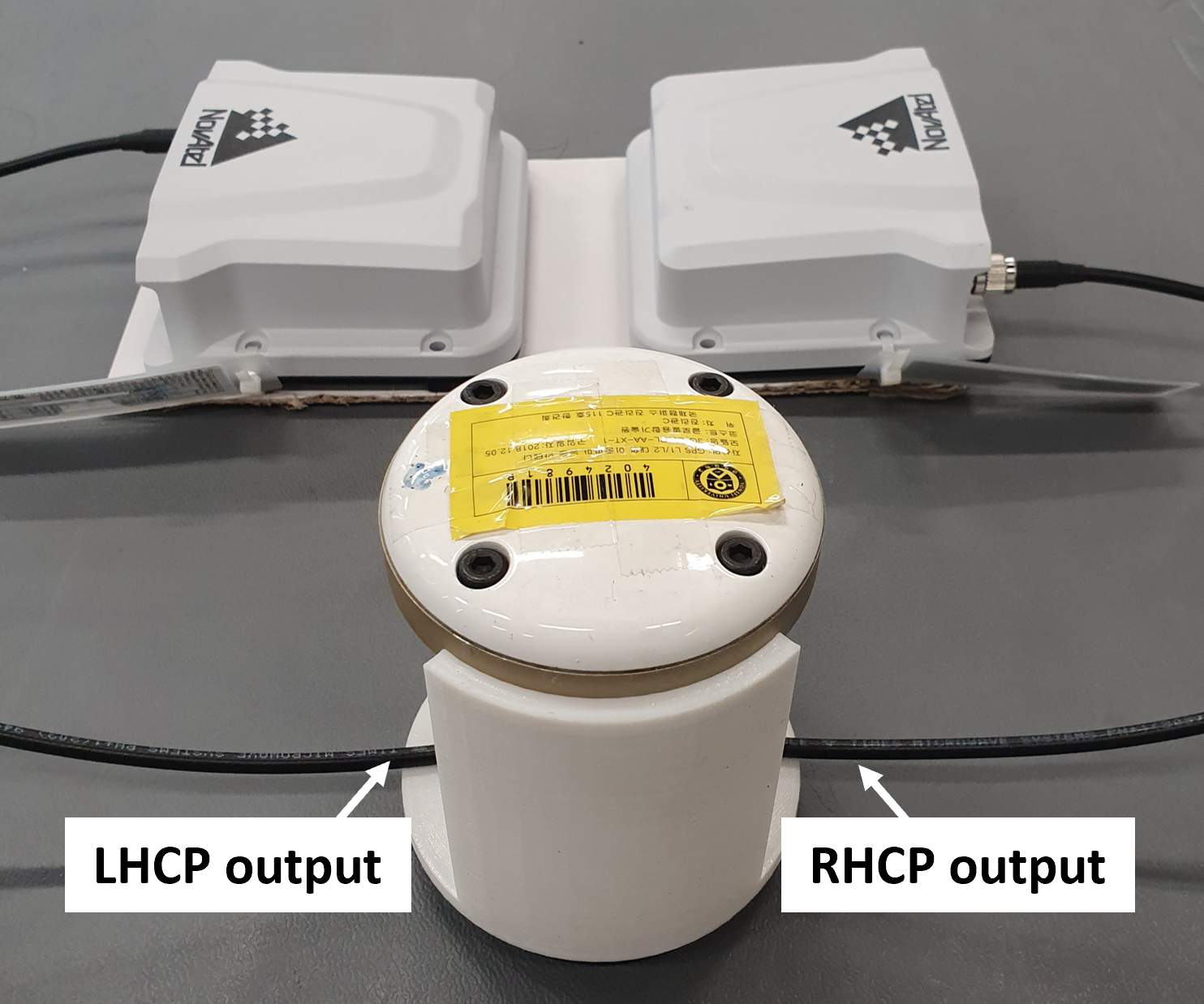}
  \caption{Dual-polarized antenna with two receivers connected.}
  \label{fig:DualPolarizedAntenna}
\end{figure}

Because the performance of a machine-learning model is significantly affected by the types of features used, it is important to extract and apply appropriate features from GNSS measurements. 
One machine-learning classification method based on a support vector machine (SVM) uses the correlation output of a GNSS signal as one of the features \cite{Suzuki17}. 
However, this method can classify the signals into only two conditions: LOS-only and multipath (i.e., NLOS-only or LOS+NLOS) conditions. 
There is a method that can classify signals into three (i.e., NLOS-only, LOS-only, and LOS+NLOS) conditions using the gradient boosting decision tree (GBDT) algorithm and three features (i.e., $C/N_0$, pseudorange residual, and satellite elevation angle) \cite{Sun19}. 
However, its classification accuracy was approximately 55\% when tested in an environment that was different from the training environment. 
Another machine-learning-based signal reception classification method compared four machine-learning algorithms, including the GBDT algorithm \cite{Kim22:Machine}. 
This method utilizes five features, including the double-difference pseudorange residual from dual single-polarized antennas, to improve the classification performance. A classification accuracy of 48\%–63\% was achieved when the test was performed under different environments from the training environment.

In this paper, we propose a machine-learning-based method to classify GPS signal reception conditions using a dual-polarized antenna and compare its performance with that of an existing method \cite{Kim22:Machine} that uses dual single-polarized antennas. 
This dual-polarized antenna solution is more beneficial than the dual single-polarized antenna solution because of its smaller form factor. 
When a dual-polarized antenna is used, both the RHCP and LHCP signals can be received, enabling the extraction of useful features for multipath detection. 
We used three features (i.e., satellite elevation angle, $C/N_0$ of RHCP signal, and $C/N_0$ difference between RHCP and LHCP signals) and the decision tree algorithm for classification.
The decision tree algorithm was chosen because it demonstrated the best performance during our tests out of the four algorithms evaluated in \cite{Kim22:Machine}.

\section{GPS Signal Collection and Labeling}

\subsection{GPS Signal Collection}
Unlike the single-polarized antenna, the dual-polarized antenna we used for GPS signal collection has both RHCP and LHCP outputs in a single antenna. As shown in Fig.~\ref{fig:DualPolarizedAntenna}, the RHCP and LHCP signals can be collected by attaching a receiver to each antenna output.

\begin{figure}
  \centering
  \includegraphics[width=0.8\linewidth]{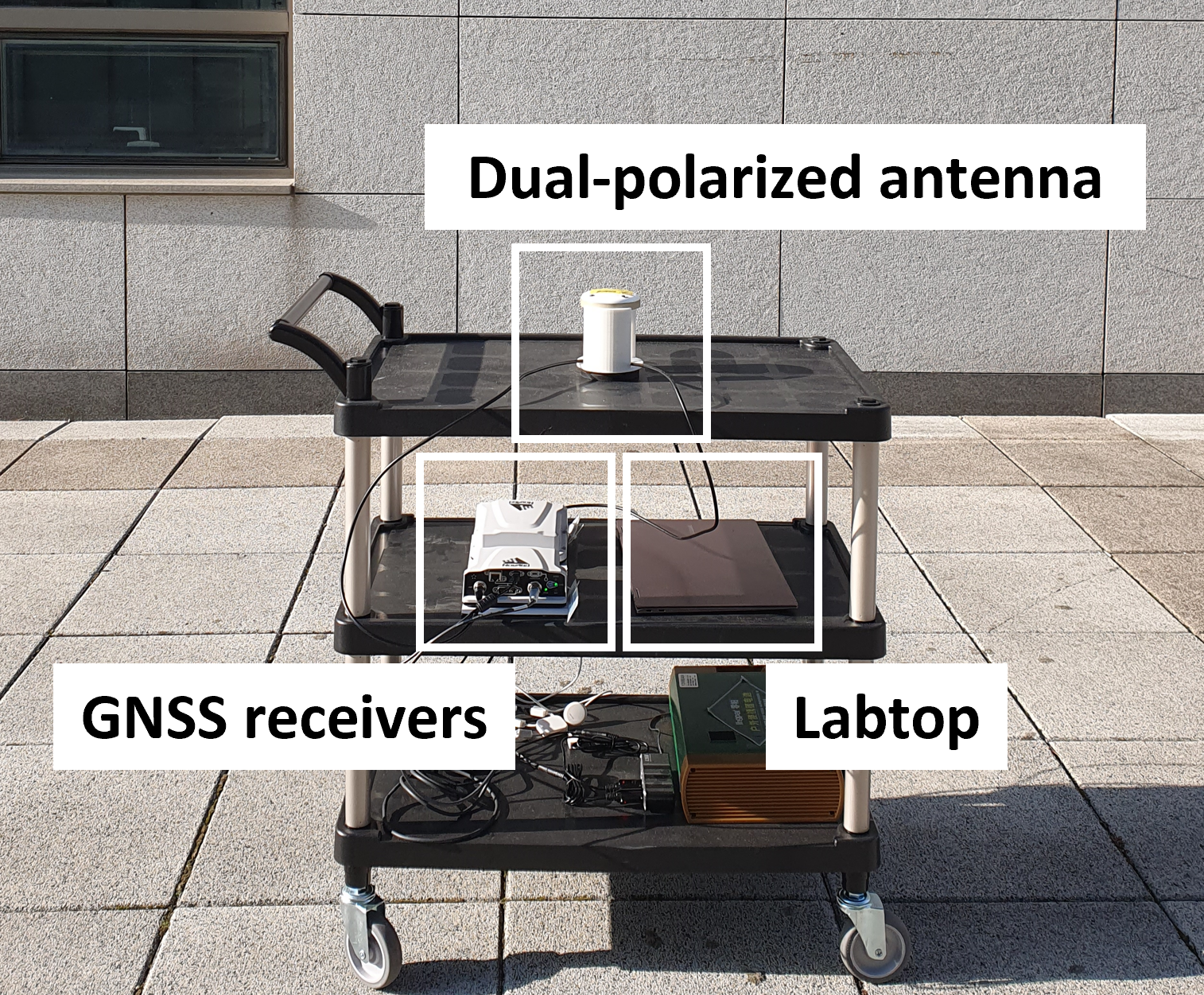}
  \caption{Hardware setup for GPS signal collection.}
  \label{fig:Hardware}
\end{figure}

\begin{table}
\centering 
\begin{threeparttable}
\centering
\caption{Numbers of data samples for each dataset}
\renewcommand{\arraystretch}{1.5}
\begin{tabular}{l||cccc}\hline
 GPS dataset & T0\tnote{1} & T1\tnote{2} & T2\tnote{3} & T3\tnote{4}\\
 \hline
 Total samples & 7500 & 7500 & 6311 & 11575 \\
 NLOS-only samples & 2500 & 2500 & 2038 & 3608 \\
 LOS-only samples & 2500 & 2500 & 2195 & 4033 \\
 LOS+NLOS samples & 2500 & 2500 & 2078 & 3934 \\
 \hline
\end{tabular}
\begin{tablenotes}
\item [1] Training dataset collected from locations A, B, and C
\item [2] Testing dataset collected from locations A, B, and C
\item [3] Testing dataset collected from location D
\item [4] Testing dataset collected from location E
\end{tablenotes}
\label{table:LabelResult}
\end{threeparttable}
\end{table}

The hardware setup shown in Fig.~\ref{fig:Hardware} was used to collect GPS L1 signals for generating machine-learning training and testing datasets. 
The hardware setup consisted of an Antcom 3G1215RL-AA-XT-1 dual-polarized antenna, two NovAtel PwrPak7 receivers, and a laptop. 
For comparison with the existing method that uses dual single-polarized antennas \cite{Kim22:Machine}, GPS signal collection was performed at the same five locations labeled as A, B, C, D, and E, all of which had similar multipath environments. 
For the dual-polarized and dual single-polarized antennas, the training and testing datasets were created in the same manner. 
The numbers of data samples for the training and testing datasets are presented in Table \ref{table:LabelResult}.

\subsection{Generation of Ground-Truth Signal Reception Condition}

Generating the ground truth signal reception condition for the collected signal, which is either NLOS-only, LOS-only, or LOS+NLOS, is necessary to prepare labeled datasets for supervised learning. 
Using a commercial 3D building map from \textit{3dbuildings} \cite{3Dbuildings} and commercial ray tracing software called \textit{Wireless InSite} \cite{WirelessInsite}, we generated a ground-truth signal reception condition for each satellite signal.
In the ray tracing simulation, we made the assumption that all 3D building surfaces were composed of concrete.
This is the same approach with \cite{Kim22:Machine}.
Finally, the labeled training and testing datasets for supervised learning were prepared by concatenating the generated ground-truth signal reception conditions with the feature vectors extracted from the signals.
The applied features will be explained in the following section.

\section{Feature Selection and Decision Tree Algorithm}

\subsection{Selected Features from Dual-Polarized Antenna Signals}

The features used in this study were the satellite elevation angle, $C/N_0$ of the RHCP signal, and $C/N_0$ difference between the RHCP and LHCP signals.
Guermah \textit{et al.} \cite{Guermah18} proposed a signal reception condition classification method that uses satellite elevation angle and $C/N_0$ difference between the RHCP and LHCP signals. 
We used additional feature (i.e., $C/N_0$ of the RHCP signal) to enhance the classification accuracy.
These features have the following characteristics.

\begin{itemize}
\item Satellite elevation angle: A low-elevation satellite is more susceptible to multipaths than a high-elevation satellite because it is more likely to be blocked or reflected by buildings.

\item $C/N_0$ of RHCP signal: $C/N_0$ is the ratio of the received carrier (i.e., signal) power to noise density, whose unit is dB-Hz. Generally, when a signal is reflected, propagation loss occurs owing to an additional propagation path \cite{Hsu17}; therefore, the LOS-only condition has a higher $C/N_0$ value than the other conditions. As shown in Fig.~\ref{fig:Feature}, $C/N_0$ of the LOS-only condition is within a range of 36--51 dB-Hz, whereas that of the NLOS-only and LOS+NLOS conditions are widely distributed in the range of 25--51 dB-Hz.

\item $C/N_0$ difference between RHCP and LHCP signals: The RHCP and LHCP antenna components are sensitive to direct signals and multipath signals, respectively. 
Therefore, the $C/N_0$ difference between the RHCP and LHCP signals is an effective parameter for detecting multipath conditions. Fig.~\ref{fig:Feature} shows that all $C/N_0$ difference values (i.e., RHCP $C/N_0$ -- LHCP $C/N_0$) in the LOS-only conditions are positive, whereas negative values are present in the NLOS-only and LOS+NLOS conditions.
\end{itemize}

\begin{figure}
  \centering
  \includegraphics[width=1.0\linewidth]{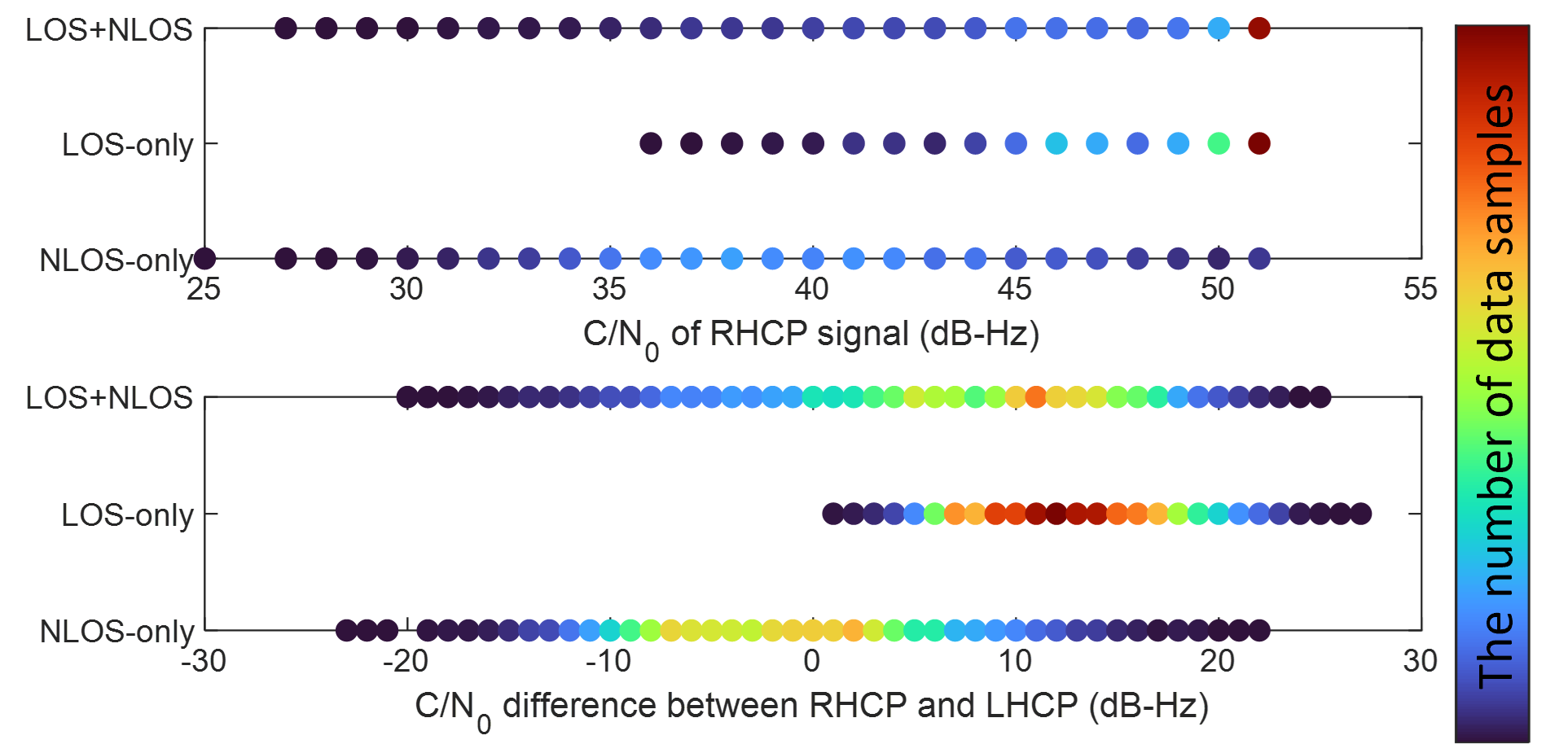}
  \caption{Distribution of $C/N_0$ of RHCP signal and $C/N_0$ difference between RHCP and LHCP signals extracted from collected data samples.}
  \label{fig:Feature}
\end{figure}

\begin{table}
\centering 
\centering
\caption{Features used for the proposed method using dual-polarized antenna compared with the existing method using dual single-polarized antennas}
\renewcommand{\arraystretch}{1.5}
\begin{tabular}{>{\centering\arraybackslash}m{4cm} >{\centering\arraybackslash}m{4cm}}\hline
\rowcolor{Gray}
Dual single-polarized antennas \cite{Kim22:Machine} & Dual-polarized antenna \\
 \hline
 Satellite elevation angle & Satellite elevation angle \\
 \hline
 $C/N_0$ & $C/N_0$ of RHCP signal \\
 \hline
 Temporal difference of $C/N_0$ & $C/N_0$ difference between RHCP and LHCP signals \\
 \hline
 Difference between delta pseudorange and pseudorange rate \\
 \hline
 Double difference pseudorange residual \\
 \hline
\end{tabular}
\label{table:ComparingFeatures}
\end{table}

Table \ref{table:ComparingFeatures} summarizes the features used in this study compared with the existing method in \cite{Kim22:Machine}. 
Instead of the five features extracted from dual single-polarized antenna signals, we used only three features extracted from dual-polarized antenna signals.

\subsection{Decision Tree Algorithm for Classification}

We used a decision tree algorithm to classify the signal reception conditions. 
The decision tree algorithm is a very robust and well-known classification and prediction method because it can simplify complex decision-making processes, allowing decision-makers to better interpret problem solutions \cite{Damanik19}. 
Various types of decision tree algorithms are available, such as iterative dichotomiser 3 (ID3), an extension of ID3 (C4.5), and classification and regression tree (CART). 
In this study, an optimized version of the CART-based decision tree algorithm provided by \textit{scikit-learn} \cite{Pedregosa11} was used. 
We also used the \textit{GridSearchCV} function included in \textit{scikit-learn} to perform the hyperparameter tuning.

\section{Results and Discussions}

Using the T0 training dataset in Table~\ref{table:LabelResult}, we trained a machine-learning model based on the decision tree algorithm. 
We then tested the performance of the model using the T1, T2, and T3 testing datasets. 
Fig.~\ref{fig:OverallResult} shows the overall classification accuracy results for each test dataset, with accuracies of 86\%, 76\%, and 58\% obtained for T1, T2, and T3, respectively. 
T1 had the highest classification accuracy because it consisted of data samples collected from the same location as the training dataset. 
On the other hand, T2 and T3 datasets were collected from different locations, which results in lower accuracy compared to T1. 
Furthermore, the accuracy of T3 was relatively lower than that of T2, which appeared to be because of the difference in the number of data samples in the two datasets. 
As shown in Table~\ref{table:LabelResult}, T3 had more data samples than T0 (i.e., training dataset) and approximately twice as many as T2. Generally, when using a testing dataset that is much larger than the training dataset, many untrained feature-domain data samples may be included, resulting in performance degradation \cite{Racz21}.

The results of the classification accuracy according to the NLOS-only, LOS-only, and LOS+NLOS conditions are shown in Fig.~\ref{fig:ClassResult}. Our trained machine-learning model classified the LOS-only condition most accurately; however, the classification accuracy of the LOS+NLOS condition was relatively low. These results are expected because the LOS+NLOS environment is more diverse than the NLOS-only or LOS-only environments. Therefore, to increase the classification accuracy of the LOS+NLOS condition, additional LOS+NLOS data samples should be used for training.

Finally, we compared the performance of the dual-polarized antenna-based machine-learning model with that of the existing method that uses dual single-polarized antennas. 
We compared two performance results: classification accuracy (T1) and classification accuracy (T2+T3), as listed in Table \ref{table:ComparingResult}. 
The former was the test result using the testing dataset T1 collected from the same location as the training dataset, and the latter was the test result using testing datasets T2 and T3 collected from different locations. 
Note that the result of classification accuracy (T2+T3) is more relevant to the expected performance in actual applications than the result of classification accuracy (T1).

As shown in Table \ref{table:ComparingResult}, classification accuracy (T1) had similar results, with approximately 86\% in both cases. 
The classification accuracy (T2+T3) of the method using a dual-polarized antenna was 64.47\%, indicating slightly better performance than the method using dual single-polarized antennas. 
Considering that the dual-polarized antenna solution has a smaller form factor, we can conclude that the proposed method is more beneficial than the existing one.

\begin{figure}
  \centering \includegraphics[width=0.95\linewidth]{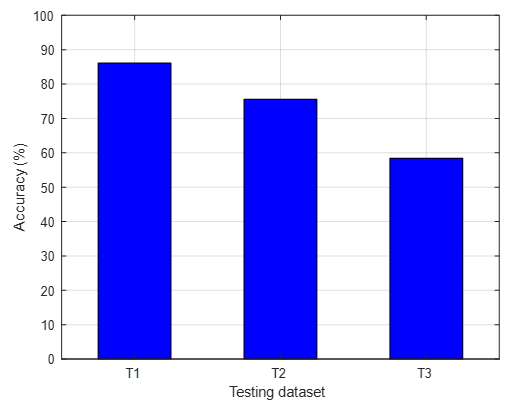}
  \caption{Overall classification accuracy for the three testing datasets.}
  \label{fig:OverallResult}
\end{figure}

\begin{figure}
  \centering \includegraphics[width=1.0\linewidth]{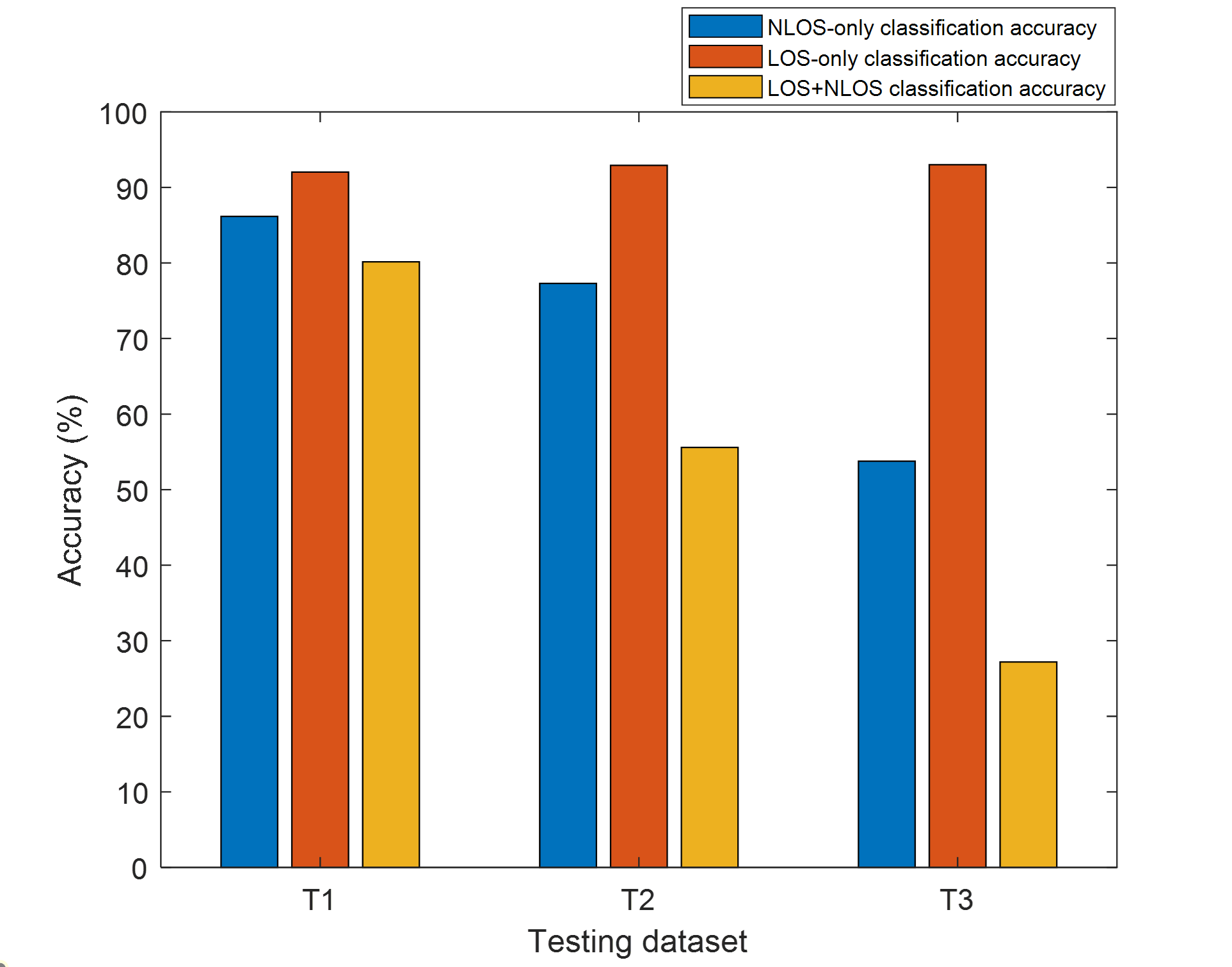}
  \caption{Classification accuracy according to the signal reception conditions for the three testing datasets.}
  \label{fig:ClassResult}
\end{figure}

\begin{table}
\centering 
\centering
\caption{Comparison of classification accuracy between dual single-polarized antennas and dual-polarized antennas}
\renewcommand{\arraystretch}{1.5}
\begin{tabular}{|>{\centering\arraybackslash}m{2.5cm}|>{\centering\arraybackslash}m{2.5cm}|
>{\centering\arraybackslash}m{2.5cm}|}\hline
 & Classification accuracy (T1) & Classification accuracy (T2+T3) \\
 \hline
 Dual single-polarized antennas & 86.81\% & 63.10\% \\
 \hline
 Dual-polarized antenna & 86.12\% & \textbf{64.47\%} \\
 \hline
\end{tabular}
\label{table:ComparingResult}
\end{table}

\section{Conclusion}

In this study, we developed a decision tree-based method of classifying GPS signal reception conditions using dual-polarized antennas and compared its performance with that of an existing method using dual single-polarized antennas. 
A test using datasets collected at different locations from the training dataset had a classification accuracy of 64.47\%, which was slightly higher than the accuracy of the existing method (63.10\%). 
The performance is expected to be further enhanced if the LOS+NLOS conditions, which exhibit low classification accuracy, are additionally collected and used for training. 
In conclusion, because of the additional advantage of having a compact form factor, the dual-polarized antenna solution is more beneficial than the dual single-polarized antenna solution.

\bibliographystyle{IEEEtran}
\bibliography{mybibfile, IUS_publications}

\end{document}